\documentclass[preprint,12pt]{elsarticle}
\usepackage{amssymb}
\usepackage{amsmath}
\usepackage{url}
\usepackage{newunicodechar}
\newunicodechar{−}{-} 
\usepackage{tabularray}
\usepackage{colortbl}
\usepackage{hhline}
\usepackage{subcaption}
\usepackage{multirow}
\usepackage{hyperref}
\journal{Advanced Engineering Informatics}
\begin{document}
\begin{frontmatter}
\title{Extracting ORR Catalyst Information for Fuel Cell from Scientific Literature}
\author[1]{Hein Htet} 
\author[1]{Amgad Ahmed Ali Ibrahim}
\author[2]{Yutaka Sasaki}
\author[1]{Ryoji Asahi}
\affiliation[1]{organization={Institutes of Innovation for Future Society, Nagoya University},
            addressline={Furo-cho, Chikusa-ku}, 
            city={Nagoya},
            postcode={464-8603}, 
            country={Japan}}
\affiliation[2]{organization={Computational Intelligence Laboratory, Toyota Technological Institute},
            addressline={2-12-1 Hisakata, Tempaku-ku}, 
            city={Nagoya},
            postcode={468-8511}, 
            country={Japan}}
\begin{abstract}
The oxygen reduction reaction (ORR) catalyst plays a critical role in enhancing fuel cell efficiency, making it a key focus in material science research.
However, extracting structured information about ORR catalysts from vast scientific literature remains a significant challenge due to the complexity and diversity of textual data. 
In this study, we propose a named entity recognition (NER) and relation extraction (RE) approach using DyGIE++ with multiple pre-trained BERT variants, including MatSciBERT and PubMedBERT, to extract ORR catalyst-related information from the scientific literature, which is compiled into a fuel cell corpus for materials informatics (FC-CoMIcs). A comprehensive dataset was constructed manually by identifying $12$ critical entities and two relationship types between pairs of the entities. Our methodology involves data annotation, integration, and fine-tuning of transformer-based models to enhance information extraction accuracy. We assess the impact of different BERT variants on extraction performance and investigate the effects of annotation consistency.
Experimental evaluations demonstrate that the fine-tuned PubMedBERT model achieves the highest NER F1-score of $82.19\%$ and the MatSciBERT model attains the best RE F1-score of $66.10\%$. Furthermore, the comparison with human annotators highlights the reliability of fine-tuned models for ORR catalyst extraction, demonstrating their potential for scalable and automated literature analysis. The results indicate that domain-specific BERT models outperform general scientific models like BlueBERT for ORR catalyst extraction.
\end{abstract}
\begin{graphicalabstract}
\includegraphics[width=14cm, height=10.0cm]{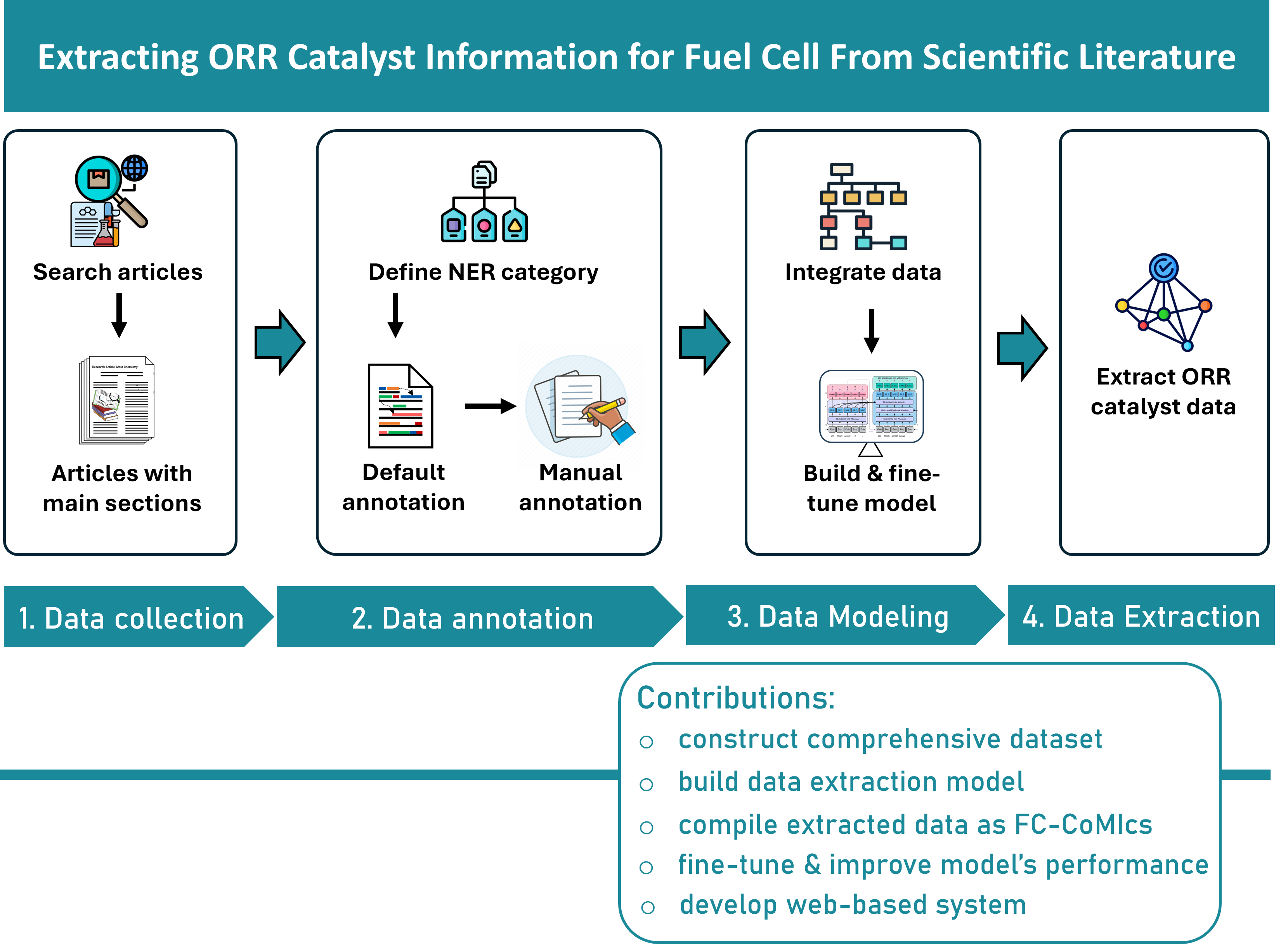}
\end{graphicalabstract}
\begin{highlights}
\item Construction of a comprehensive dataset with 12 entities and 2 relationships.
\item Automated information extraction for  oxygen reduction reaction (ORR) catalysts in  polymer electrolyte fuel cell (PEFC) using NLP techniques.
\item Extracted data is compiled as fuel cell corpus for materials informatics (FC-CoMIcs).
\item Enhanced entity and relation recognition by fine-tuning pre-trained BERT models within the DyGIE++ framework.
\item Developed a web-based system to facilitate data collection, annotation, and extraction.
\end{highlights}
\begin{keyword}
Natural Language Processing (NLP) \sep Information Extraction \sep Fuel Cell Catalysts\sep Oxygen Reduction Reaction (ORR)
\end{keyword}
\end{frontmatter}

\section{Introduction}
Polymer Electrolyte Fuel Cell (PEFC) is a key component of clean energy technology, offering a high energy efficiency, a low greenhouse gas emission, and scalability for various applications including automotive and stationary power generation \cite{Wang2011}. A critical factor influencing PEFC performance is the Oxygen Reduction Reaction (ORR), which takes place at the cathode and significantly influences the overall cell performance. Although ORR significantly influences the performance of various fuel cells, such as Molten Carbonate Fuel Cells (MCFCs), Phosphoric Acid Fuel Cells (PAFCs), and Solid Oxide Fuel Cells (SOFCs), this study focuses on ORR catalysts for PEFCs, where catalyst performance plays a crucial role in achieving high energy efficiency and commercial viability. The development of advanced catalysts for ORR has been an active area of research, with a focus on improving catalytic activity, durability, and cost-effectiveness. Advanced ORR catalysts, typically composed of precious metals like platinum or their alloys, have been developed to enhance catalytic activity and stability while reducing material costs \cite{Wu2013}. The continual advancement of ORR catalysts is critical for overcoming challenges in PEFC commercialization and addressing the increasing demand for sustainable energy solutions worldwide.

In addition to traditional noble metal-based catalysts, recent research has explored non-precious metal catalysts, doped carbons, and single-atom catalysts (SACs) as viable alternatives \cite{Cui2021}. These efforts aim to make a balance between cost-effectiveness and high catalytic efficiency. However, the effectiveness of these materials is influenced by various factors, such as the synthesis process, support material, and operating conditions, further emphasizing the complexity of ORR catalyst development.

The rapid expansion of scientific literature on fuel cells and ORR catalysts poses significant challenges for researchers. Key information related to catalyst composition, synthesis methods, and performance metrics is often scattered across diverse publications in chemistry, materials science, and engineering domains. Traditional manual approaches to literature review are time consuming, labor-intensive, and may overlook critical data, making it difficult to extract key insights effectively with the growing body of knowledge. Moreover, critical relationships between entities, such as the impact of synthesis conditions on catalytic activity, may remain unexplored due to the limitations of manual analysis. As such, there is a growing need for automated tools to facilitate the extraction of structured information, enabling researchers to focus on analysis and discovery \cite{Kononova2021}.

This study aims to bridge the gap in literature mining by developing a framework for information extraction specific to ORR catalyst research for PEFCs. As the primary data source, we focused on collecting literature in HTML format from the \emph{Royal Society of Chemistry (RSC)} to ensure access to high-quality and relevant publications. Our approach combines manual annotation with advanced machine learning techniques to construct a high-quality, domain-specific dataset and train a model capable of extracting meaningful information from scientific literature. 

The key contributions of this study are as follows:
\begin{enumerate}
\item Development of a web-based system that integrates data collection, annotation, and extraction features, providing an accessible platform for researchers to explore and analyze ORR catalyst data.
\item Construction of a comprehensive dataset using the Brat annotation tool, identifying $12$ critical entities such as catalysts, supports, and values, along with two relationship types, i.e., equivalent and related\_to.
\item Fine-tuning of the DyGIE framework with pre-trained BERT models, including SciBERT, MatSciBERT, and PubMedBERT, to perform precise NER and RE.
\item Evaluation of the framework’s performance in extracting complex material science concepts and their interrelationships, demonstrating its potential to accelerate catalyst discovery.
\end{enumerate}

By automating the labor-intensive process of literature mining, our framework not only reduces the time and effort required for data extraction but also enhances the reliability and reproducibility of research findings. Extracted data is compiled as fuel cell corpus for materials informatics (FC-CoMIcs). The insights gained from this study can serve as a foundation for further advancements in ORR catalyst design, synthesis optimization, and performance evaluation. To facilitate the adoption of our framework, we have developed a web-based system that enables researchers to access, annotate, and extract information in a user-friendly interface. This system supports collaborative annotation and allows for real-time exploration of extracted data. Future enhancements to the system include the integration of predictive modeling features, enabling researchers to forecast catalyst performance based on extracted properties and relationships.

The rest of this paper is organized as follows: Section~\ref{literature_review} provides a review of related studies. Section~\ref{methodology} details the proposed methodology, while Section~\ref{result_and_discussion} presents and discusses the results. Finally, Section~\ref{conclusion} summarizes the findings and outlines directions for future work.

\section{Related Studies}
\label{literature_review}
In this section, we review the literature related to the topics discussed in our paper. We organize the section into \emph{three} main themes: Material Science Information Extraction, NER and RE Techniques, and Applications of NLP in Catalyst Research.

\subsection{Material Science Information Extraction}
Information extraction (IE) in material science has become an increasingly important tool for processing and analyzing large amounts of scientific data. 

Kyosuke Yamaguchi et al. \cite{Yamaguchi2022} constructed a superconductivity corpus, SC-CoMIcs, consisting of 1,000 manually annotated abstracts for NER, RE, and main material identification. Using DyGIE++ and Longformer-based models, they achieved high F1 scores ranging from 73–97\% across tasks, enabling automatic extraction of key superconductivity-related information such as transition temperature, dopant, and site. Notably, the extracted doping data was found to align with the Hume–Rothery rules, highlighting the corpus’s potential for discovering physical-chemical insights. This work demonstrates the value of domain-specific corpora in advancing materials informatics through literature mining.

Rui Zhang et al. \cite{Zhang2023} developed a method to extract textual and tabular data from materials science literature, using a SciBERT-based model for NER and a table recognition method for material compositions. Applied to over $11,000$ papers on stainless steel, their approach extracted millions of entities and predicted material property trends using Gradient Boosting Decision Tree models, emphasizing the value of integrating textual and tabular data for literature mining.

Mara Schilling-Wilhelmi et al. \cite{Schilling-Wilhelmi2025} reviewed the potential of large language models (LLMs) for extracting structured data from unstructured chemical literature. They emphasized LLMs as scalable solutions compared to traditional and partially automated methods, enabling efficient extraction of actionable data. The review also highlighted the importance of integrating domain expertise with LLMs, showcasing their potential to accelerate data-driven discovery in chemistry and materials science.

Wei Zhang et al. \cite{Zhang2024} demonstrated the effectiveness of fine-tuned LLMs for chemical text mining tasks such as entity recognition, reaction role labeling, and MOF synthesis extraction. Fine-tuned ChatGPT models achieved high accuracy ($69\%$–$95\%$) with minimal annotated data, outperforming models trained on larger datasets. The study highlights fine-tuned LLMs as versatile tools for automated chemical knowledge extraction, advancing data-driven innovations in chemistry.

Ankan Mullick et al. \cite{Mullick2024} proposed MatSciRE, a Pointer Network-based encoder-decoder framework, for joint entity and relation extraction in material science literature. Focused on battery materials, the model extracts triplets (entity1, relation, entity2) for relations like conductivity, capacity, and voltage. MatSciRE achieved an F1-score of $0.771$, outperforming ChemDataExtractor (CDE) ($0.716$). The study highlights its utility with a curated dataset, and a web-based API for extracting triplets from manuscripts, contributing to the development of material science knowledge bases.

\subsection{NER and RE Techniques}
NER and RE are widely used in NLP for structuring unstructured data by identifying key entities and their relationships. 

John Dagdelen et al. \cite{Dagdelen2024} proposed a method using fine-tuned LLMs (GPT-3, Llama-2) for NER and RE in scientific texts, focusing on materials chemistry. The approach efficiently extracts and structures complex data with minimal training, and technical expertise, outperforming traditional models. While challenges like schema formatting and hallucinations exist, the study demonstrates the potential of LLMs as accessible tools for enhancing NER and RE tasks in scientific data extraction. 

L. Weston et al. \cite{Weston2019} applied NER to extract structured data from over $3.27$ million materials science abstracts, achieving $87\%$ accuracy (F1-score) on this dataset and extracting over $80$ million entities. This method enables efficient information retrieval and the answering of complex “meta-questions” using simple database queries. While the system currently analyzes abstracts,  the approach can be expanded to full texts, advancing the use of NER and RE for materials science data extraction.

Nandita Goyal and Navdeep Singh \cite{Goyal2025} provided a comprehensive survey of NER and RE techniques in biomedical literature. They highlighted advancements, such as RNN-CRF, Word2Vec, and BERT-based models, while noting challenges like handling domain-specific nuances and joint modeling complexities. Future directions include adapting pre-trained models like BioBERT, improving data quality through a data-centric approach, exploring advanced neural architectures, and integrating external knowledge to enhance performance.

\subsection{Applications of NLP in Catalyst Research}
The use of NLP in catalyst research is a growing area of interest, as it offers the potential for automating literature mining to extract key information about catalysts, their performance, and synthesis methods. 

Manu Suvarna et al. \cite{Suvarna2023} introduced a transformer model for extracting synthesis protocols in heterogeneous catalysis, demonstrated using SACs. The model converts protocols into action sequences, enabling statistical inference of synthesis trends and applications across catalyst families. This work underscores the importance of collaboration between catalysis researchers and machine learning experts to advance automated synthesis and data-driven catalyst discovery.

Yuming Su et al. \cite{Su2024} discussed the transformative impact of LLMs and AI technologies on catalyst design and discovery, shifting from traditional trial-and-error approaches to high-throughput, data-driven methodologies. They highlighted advancements in automated information extraction, robotic experimentation, active machine learning, and interpretable models for catalyst design. Despite challenges in multimodal data integration and system standardization, the review emphasized LLMs' potential in comprehending complex data and supporting autonomous, intelligent systems in catalyst research. The integration of these AI technologies promises to accelerate the exploration of chemical spaces and enhance scientific discovery.

Sukriti Singh and Raghavan B. Sunoj \cite{Singh2022} developed a transfer learning (TL) protocol using SMILES-based representations to predict yields and enantioselectivities in homogeneous catalysis. Trained on over $1$ million molecules, the model achieved high accuracy across three reaction types, including Buchwald–Hartwig cross-coupling (RMSE: $4.89$), enantioselective N,S-acetal formation (RMSE: $8.65$), and asymmetric hydrogenation (RMSE: $8.38$). Approximately $90$–$97$\% of the predicted yield and enantiomeric excess were within 10 units of the actual experimental values, demonstrating the TL model's efficiency and applicability to diverse data sizes. This approach highlights the potential of TL in high-throughput reaction discovery and sustainable catalysis.

\section{Methodology}
\label{methodology}
\begin{figure}[!ht]
\centering
\includegraphics[width=11cm, height=9.0cm]{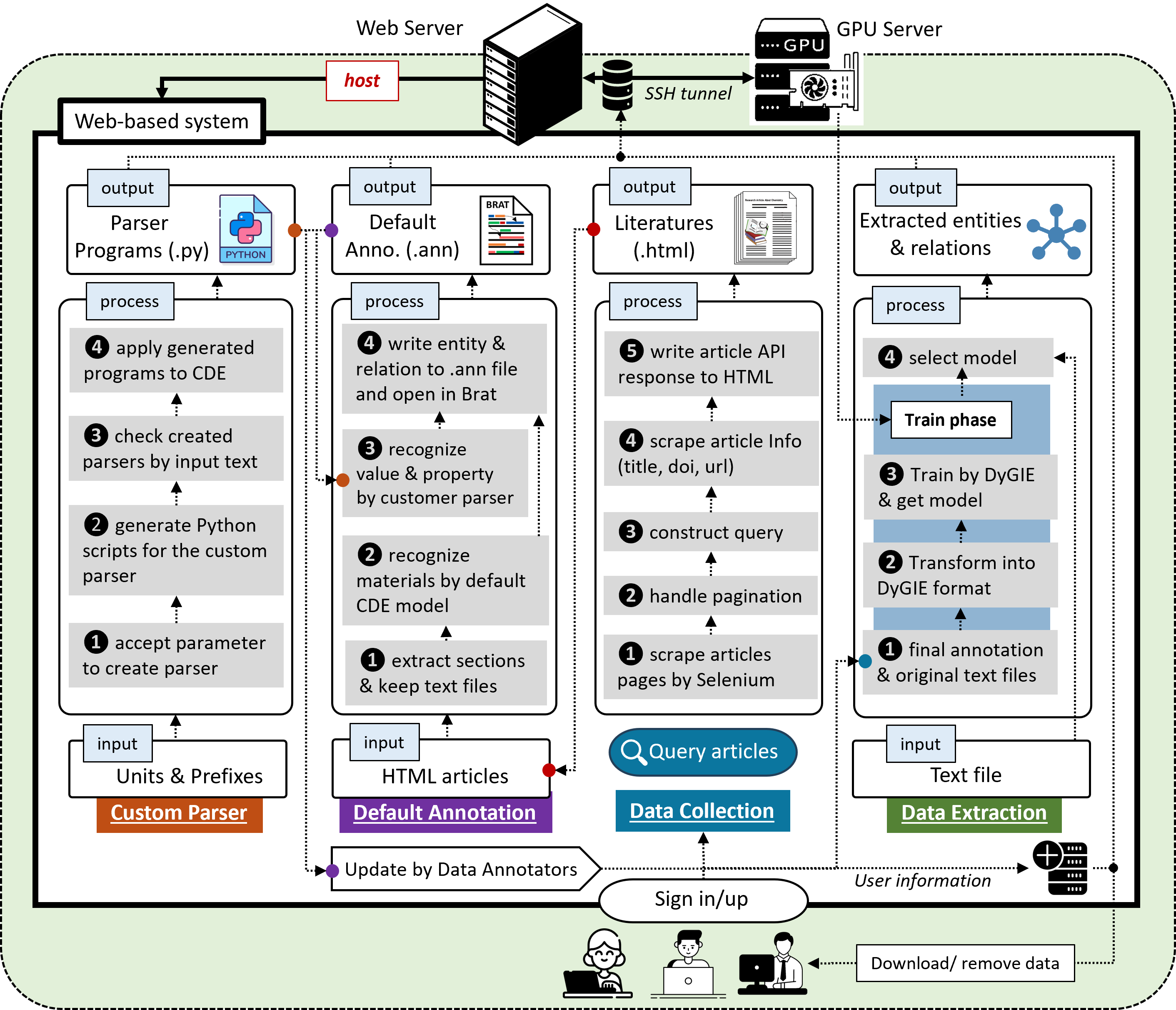}
\caption{System overview: web-based ORR catalyst data collection and analysis.}
\label{fig:system_overview}
\end{figure}

In this section, we outline the methodology employed to collect, annotate, integrate, and model data related to ORR catalysts for fuel cells. The process is divided into \emph{four} main stages: \emph{data collection}, \emph{data annotation}, \emph{data integration \& modeling}, and \emph{data extraction}. Each stage involves the use of specific tools and techniques aimed at ensuring the accuracy and efficiency of the data processing pipeline. To facilitate this workflow, we have developed a web-based system platform, as shown in Figure~\ref{fig:system_overview}, allowing researchers to perform all these stages seamlessly through the platform. The tools and methods employed for selecting fuel cell related articles, annotating key entities and relationships, preparing datasets, and building models are described in detail, followed by the information extraction with the annotation models.

\vspace{-1ex}

\subsection{Data Collection}
The success of the information extraction process relies heavily on the quality and diversity of the dataset. In this research, an automated data collection approach was employed to gather relevant articles for training and evaluation. Through the developed web-based platform, researchers can efficiently search and collect a large volume of scientific literature by specifying their queries.

\subsubsection{Adopted Tools \& Techniques}
To collect the data necessary for this study, we utilized the following tools and techniques:  
\vspace{-1ex}
\begin{itemize}
\setlength{\itemsep}{1.6pt}
\setlength{\parskip}{1.6pt}
    \item \textbf{Article Database}: It serves as a comprehensive repository \cite{Antony2014} of scientific literature, particularly focused on chemistry and materials science, and RSC is used as a primary source for article retrieval.
    \item \textbf{Article Scraper Module}: CDE \cite{Mavračić2021} includes an RSC scraper module designed to efficiently extract relevant articles from the database, ensuring high-quality data collection.
    \item \textbf{Flask Web Framework}: Flask \cite{Miguel2018} is a lightweight web framework used to develop the web-based platform. Our web-based system was developed by integrating it with the NLP utility libraries.
\end{itemize}

\subsubsection{Fuel Cell Related Articles Selection}
For the selection of relevant articles, we used the query: \emph{ORR AND Catalyst AND (ECSA OR “mass activity” OR “ORR activity” OR “surface activity”)} to search for literature. Here, ECSA represents Electro Chemical Surface Area. The collected articles covered a wide range of publications, including journals, conference proceedings, and technical reports, published between $2010$ and $2024$, ensuring the inclusion of the most up-to-date research in the field. A total of $1259$ articles were identified. From the full-text articles, we focused on the Abstract, Results \& Discussion, and Conclusions sections, as these sections contain the most essential information.  

\begin{figure}[ht!]
\centering 
\includegraphics[width=9cm, height=6cm]{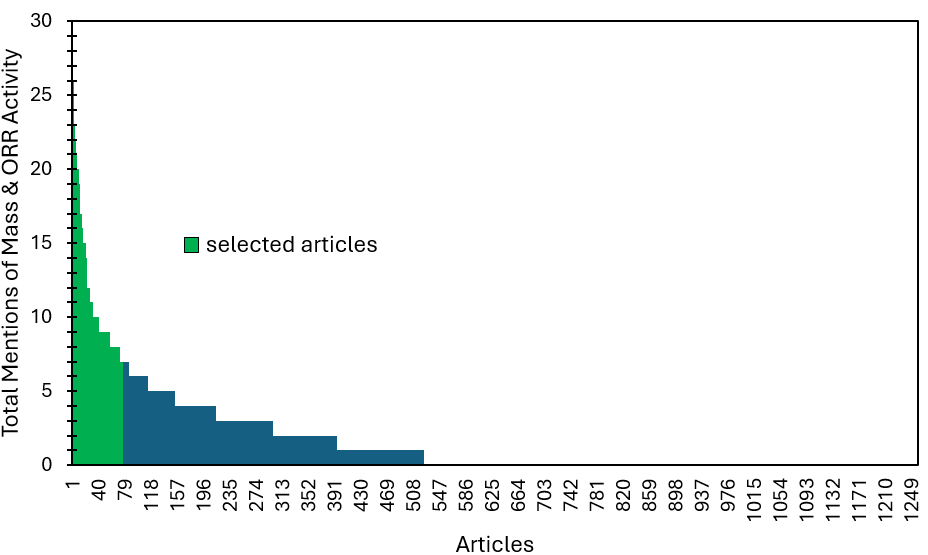}
\caption{Selected articles for data annotation.}
\label{fig:collected_annotated_data}
\end{figure}

For the present study, the articles were then ranked based on the highest number of mentions of \emph{mass activity} and \emph{ORR activity}. This process resulted in the selection of $76$ high-quality articles, as shown in Figure~\ref{fig:collected_annotated_data}, which were considered representative of the state-of-the-art in ORR catalyst research for fuel cells, and were subsequently used for data annotation.   

\subsection{Data Annotation}
Data annotation is a crucial step in the supervised learning process for NER and RE models. High-quality annotations ensure that models accurately identify and extract relevant information from unstructured text.

\subsubsection{Adopted Tools \& Techniques}
To perform the data annotation for this study, we utilized the following tools and techniques:  
\begin{itemize}
    \item \textbf{CDE's Default Parser}: CDE includes a default parser capable of recognizing chemical compositions, compounds, and material elements. Additionally, CDE allows for integrating custom parsers to fit to the present study as described below.
    \item \textbf{Custom Parser}: The developed web platform provides an interface for easily building custom parsers by specifying target units and prefixes. For example, current density can be expressed in various units (such as A cm$^{-2}$, mA cm$^{-2}$, etc.) and may include different prefixes, such as “density of current” or “current density of”. Once defined, Python scripts for the custom parsers are automatically generated and integrated into CDE’s working directory. These scripts can be downloaded and used in CDE, even when set up in another environment.
    \item \textbf{Brat}: A web-based annotation tool designed for structured text annotation \cite{Stenetorp2012}. Brat enables annotators to add structured notes to text documents, facilitating systematic data labeling that can be automatically processed and interpreted by machine learning models.
\end{itemize}

\subsubsection{Entity and Relationship Definition}
\begin{figure}[!ht]
\centering
\includegraphics[width=10.2cm, height=5.6cm]{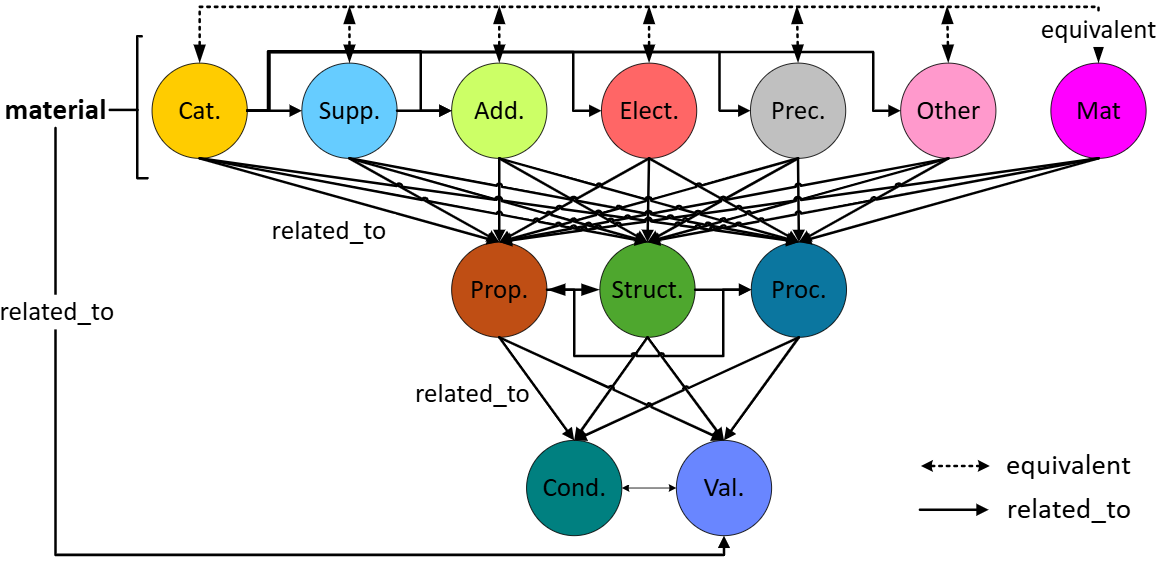}
\caption{Entity-Relation map.}
\label{fig:entity_relation}
\end{figure}

The entity types identified during annotation were based on the common components involved in ORR catalyst research. The $12$ key entities were defined as described in Table~\ref{tab:key_entities}. The identified entities are inter-related to each other. These relations are essential to extract information in a structured format such as CSV. For the present study, we focused on the two kinds of relation as defined in Table~\ref{tab:key_relations}: “related\_to” to connect between two entities as shown with arrows in Figure~\ref{fig:entity_relation}, and “equivalent”, where a material refers to a “Mat.” entity as its equivalent material.

\begin{table}[!ht]
\centering
\captionsetup{skip=3pt} 
\caption{Entities and Descriptions.}
\label{tab:key_entities}
\resizebox{0.90\textwidth}{!}{ 
\begin{tblr}{
  cell{1}{2} = {c},
  hline{1-2,14} = {-}{},
}
\textbf{No.} & \textbf{Entity (abbreviation)} & \textbf{Descriptions}                                                                                          \\
1            & Catalyst (Cat.)                & {material that enhances reaction rate\\ e.g. PtCo, CoPt3, Pt/Al2O3}                                            \\
2            & Support (Supp.)                & {material that stabilizes and disperses the catalyst\\ e.g. carbon support, Al2O3, SiO2}                       \\
3            & Additive (Add.)                & {material added to modify catalyst properties\\ e.g. N-doping, dopant}                                         \\
4            & Electrolyte (Elect.)           & {substance that conducts electricity by the \\ movement of ions\\ e.g. proton exchange membrane, Nafion}       \\
5            & Precursors (Prec.)             & {material used to synthesize the catalyst or support\\ e.g. Pt(acac)2, PtCl2, Vulcan carbon}                   \\
6            & {Other Material \\(Other)}     & {materials that are not classified as the \\above entities\\ e.g. oxygen-hydrogen, gold, silver}     \\
7            & {Material Reference\\ (Mat.)}  & {word references to the materials\\ e.g. the catalyst, the support}                                            \\
8            & Property (Prop.)               & {measurable characteristics defining material\\ performance or behavior\\ e.g. current density, mass activity} \\
9            & Structure (Struct.)            & {Atomic or molecular arrangement\\ e.g. nanoparticles, film, alloy, nanosheet}                                 \\
10           & Process (Proc.)                & {method or technique applied in synthesis\\ e.g. synthesis, sintering, sputtering, sol-gel}                    \\
11           & Condition (Cond.)              & {experimental parameters affecting reaction \\ or material behavior\\ e.g. at 700 K, (200) crystalline planes} \\
12           & Value (Val.)                   & {numerical values with units and degree of change \\in property\\ e.g. 1246.8 m2 g-1, 1600 rpm, decrease}      
\end{tblr}
}
\end{table}

\begin{table}[!ht]
\centering
\captionsetup{skip=3pt} 
\caption{Relations and Descriptions.}
\label{tab:key_relations}
\resizebox{0.90\textwidth}{!}{ 
\begin{tblr}{
  hline{1-2,4} = {-}{},
}
\textbf{No.} & \textbf{Relation} & \textbf{Description}                                                                                                    \\
1            & equivalent        & {links entities that represent the same concept or material\\ e.g. CoPt3 (equivalent) the catalyst}                     \\
2            & related\_to       & {captures connections between entities that share a \\direct association\\ e.g. PtCo (related\_to) carbon support} 
\end{tblr}
}
\end{table}

\subsubsection{Data Annotation Process}
At the beginning of the annotation process, three experts in our group created a benchmark dataset, referred to as the \emph{Gold Standard}. The accuracy and consistency of the annotators were evaluated by comparing their annotations against this \emph{Gold Standard}. The annotation process consists of two main steps:

\begin{enumerate}
    \item \textbf{Generate Default Annotations}: Initial annotations in Brat format are automatically generated using a combination of \emph{CDE’s parser} and \emph{custom-created parsers}. Figure~\ref{fig:default_annotation} shows the generated default annotation results upon the data and the colors correspond to those used for entities in Figure~\ref{fig:entity_relation}. However, the default annotations generated by CDE are not capable of identifying all the defined entities and do not produce relationships between entities. Therefore, manual annotation remains necessary to accurately extract both entities and relationships.
        \vspace{-0.5ex}
        \begin{figure}[!ht]
        \centering 
        \includegraphics[width=12cm, height=3.7cm]{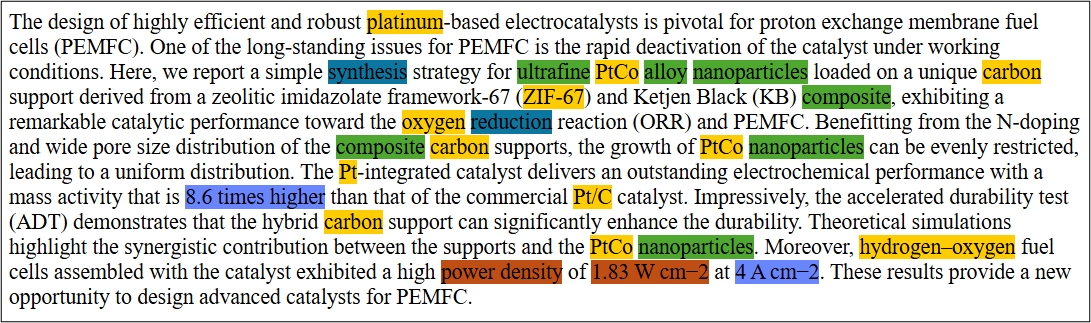}%
        \vspace{-1ex}
        \caption{An example of default annotations upon a text data taken from \cite{Zhang2024-1}.}
        \label{fig:default_annotation}
        \end{figure}

    \item \textbf{Refine Annotations}: Here, we employed three annotators to refine and perform manual annotation. Data annotators then refine and complete the annotations using the \emph{Brat annotation tool}, as shown in Figure~\ref{fig:refine_annotation}, which is hosted on our server. This setup eliminates the need for annotators to install Brat locally, providing easy access through the developed web platform.
        \begin{figure}[!ht]
        \centering
        \includegraphics[width=12cm, height=9.5cm]{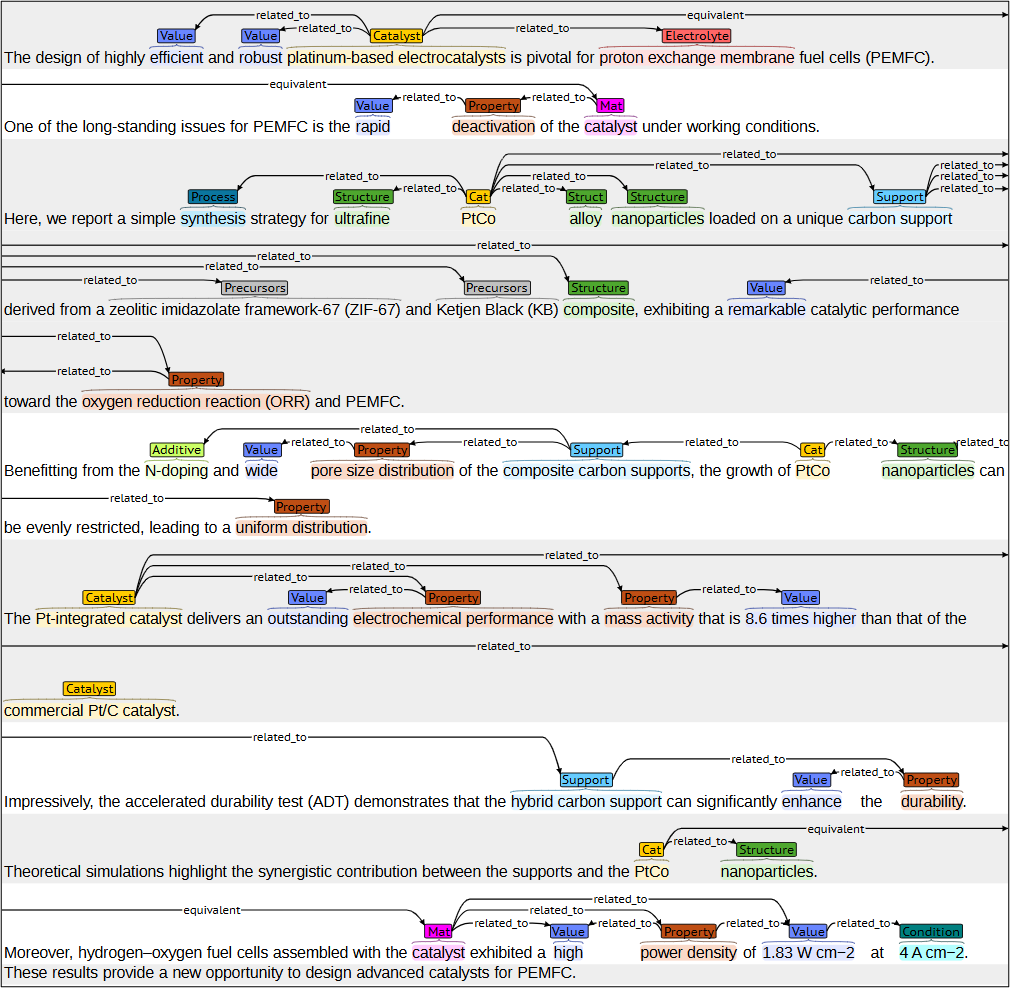}
        \caption{An example of refine default annotations on Brat by annotator upon a text data taken from \cite{Zhang2024-1}.}
        \label{fig:refine_annotation}
        \end{figure}
\end{enumerate}

\subsection{Data Integration \& Modeling}
\label{sec:data_integration}
Following the data annotation process, the next step was to integrate the annotated data and develop machine learning models for NER and RE \cite{Mitsui2023}. This phase aimed to transform the structured annotations into a format suitable for training models, ensuring high accuracy in extracting key information from ORR catalyst literature.

\subsubsection{Data Integration Process}
\begin{figure}[!ht]
\centering
\includegraphics[width=10cm, height=5cm]{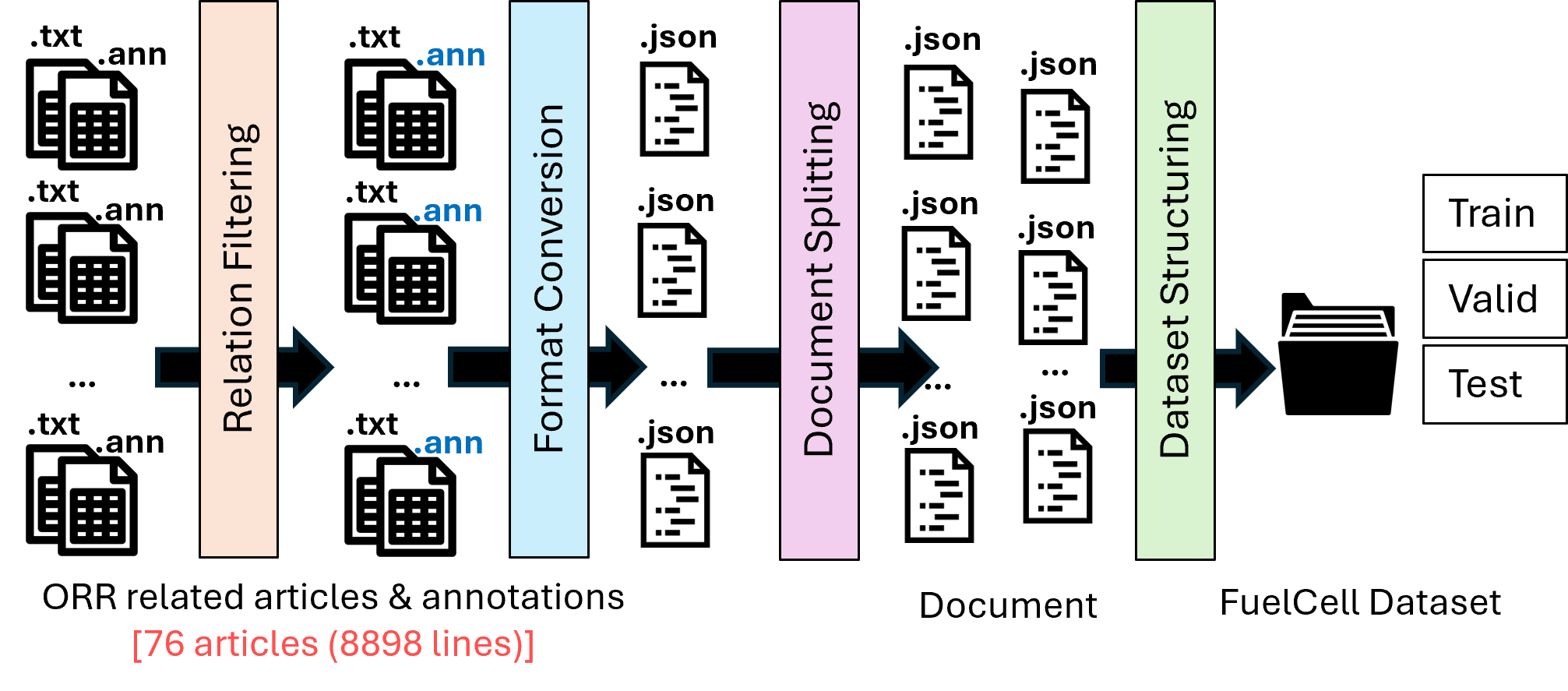}%
\caption{Fuel cell data integration process.}
\label{fig:data_integration}
\end{figure}

The integration process involved converting annotated text into a structured machine-readable format suitable for model training, as shown in Figure~\ref{fig:data_integration}. This included:

\begin{enumerate}
    \item \textbf{Relation Filtering}: Removing invalid cross-relations to ensure tokenization consistency
    \item \textbf{Format Conversion}: Converting Brat annotations into JSON format compatible with DyGIE++
    \item \textbf{Document Splitting}: Dividing documents into smaller segments to prevent CUDA out-of-memory errors
    \item \textbf{Dataset Structuring}: Organizing data into training, validation, and test sets for model evaluation
\end{enumerate}

\subsubsection{Data Modeling Process}
\label{sec:modeling_process}
The data modeling process focuses on training and fine-tuning machine learning models to extract named entities and relationships effectively from the annotated dataset. 

The key steps in this process are as follows:
\begin{enumerate}
    \item \textbf{DyGIE++ Framework}: We used the DyGIE++ framework, a graph-based model designed for NER and RE \cite{Luan2019}. DyGIE++ is particularly effective at handling complex relationships and interactions between entities in scientific text, making it suitable for extracting information from ORR catalyst literature.
    \item \textbf{Fine-tuning Pre-trained BERT Models}: In addition to DyGIE++, we fine-tuned several pre-trained BERT-based models on our annotated dataset. These included domain-specific models such as SciBERT \cite{Beltagy2019}, MatSciBERT \cite{Gupta2022}, PubMedBERT \cite{Gu2021}, and BlueBERT \cite{Peng2019}. Fine-tuning these models allowed them to adapt to the domain-specific language patterns used in ORR catalyst literature, improving their extraction accuracy.
    \item \textbf{Model Evaluation}: The performance of the fine-tuned models was evaluated using standard metrics such as Precision, Recall, and F1-score. These metrics assess the accuracy of the model predictions in extracting correct entities and relationships from the text:
\end{enumerate} 

\begin{equation}
    \text{Precision} = \frac{TP}{TP + FP}
\end{equation}

\begin{equation}
    \text{Recall} = \frac{TP}{TP + FN}
\end{equation}

\begin{equation}
    \text{F1-score} = \frac{2 \times \text{Precision} \times \text{Recall}}{\text{Precision} + \text{Recall}}
\end{equation}
where:

\begin{itemize}
\setlength{\itemsep}{1pt}
\setlength{\parskip}{1pt}
    \item $TP$ = Number of true positives (correct annotations/extractions)
    \item $FP$ = Number of false positives (incorrect annotations/extractions)
    \item $FN$ = Number of false negatives (missed annotations/extractions)
\end{itemize}

\subsection{Data Extraction}
Data extraction involves applying trained models to identify and retrieve relevant information from fuel-cell related scientific literature. The extracted data is then structured and visualized for further analysis.

\subsubsection{Model Selection \& Input Data}
\begin{figure}[!ht]
\centering 
\includegraphics[width=9.5cm, height=7.5cm]{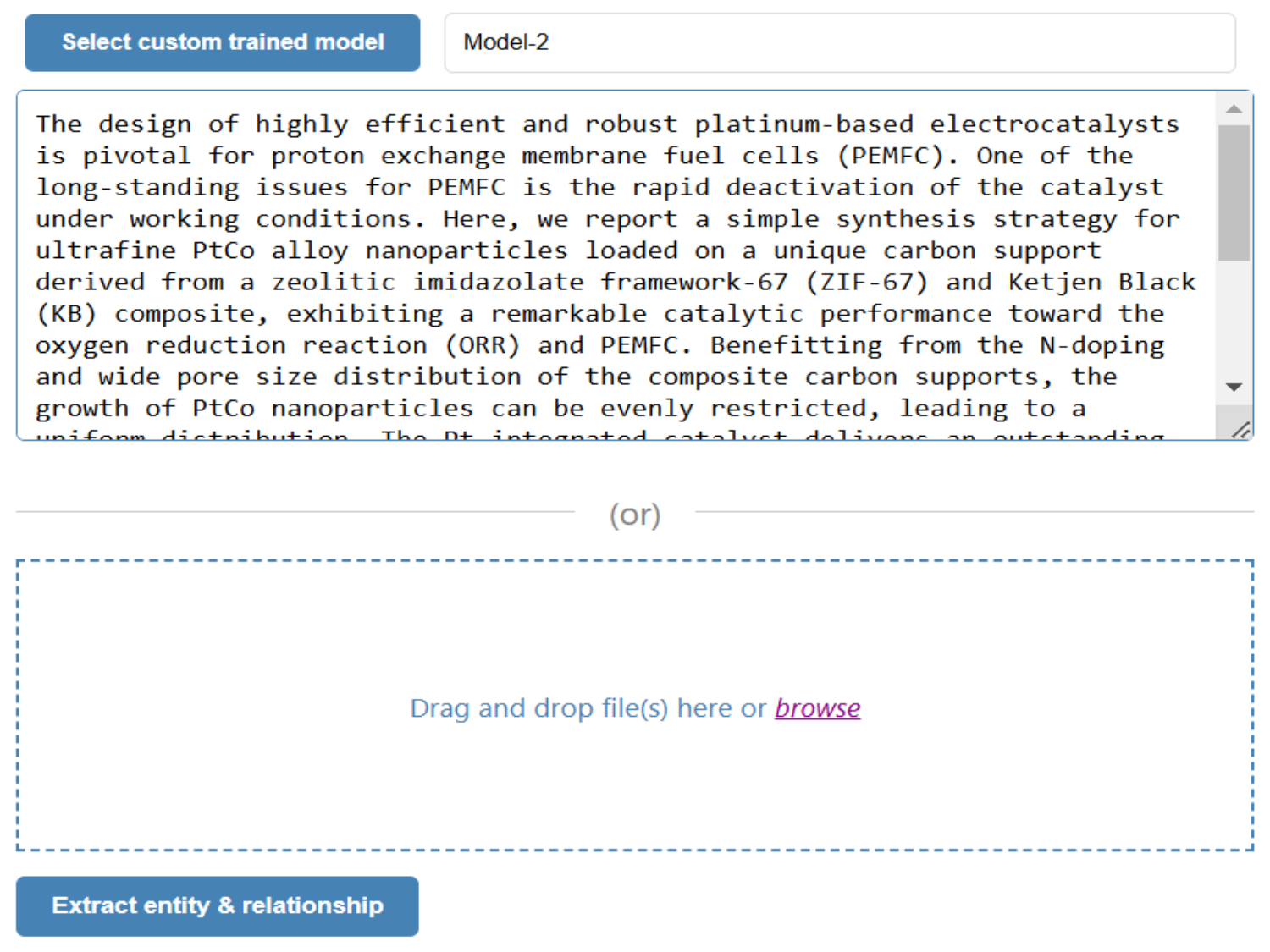}%
\caption{Model selection \& input data for extraction. The text data is taken from \cite{Zhang2024-1} as an input example.}
\label{fig:model_selection}
\end{figure}

A total of seven models were trained using various pre-trained BERT architectures. These models were employed to extract information from fuel cell related scientific literature. Users can efficiently select the preferred trained model and input data through the developed web platform to initiate the extraction process, as shown in Figure~\ref{fig:model_selection}.

\subsubsection{Extracted Output Data}
The extracted data is presented in \emph{three} formats on the web platform to facilitate interpretation and analysis:

\begin{enumerate}
    \item \textbf{Console Output} (Table~\ref{tab:ner_re_console_output}): Displays raw extracted text categorized by defined entities and relationships, providing a direct textual representation of the extracted information
        \begin{table}[!ht]
        \centering 
        \caption{Extracted data in console output.}
        \label{tab:ner_re_console_output}
        \begin{tabular}{l l}
        \hline
        \multicolumn{1}{c}{\textbf{NER}} & \multicolumn{1}{c}{\textbf{RE}} \\ \hline
        \resizebox{0.45\textwidth}{!}{ 
        \begin{tabular}[c]{@{}l@{}}
        1 platinum-based electrocatalysts (catalyst) \\
        2 proton exchange membrane (electrolyte) \\
        3 simple synthesis strategy (process) \\
        4 ultrafine (structure) \\
        5 PtCo (catalyst) \\
        6 alloy (structure) \\
        7 nanoparticles (structure) \\
        8 carbon support (support) \\
        9 Ketjen Black (KB) (support) \\
        10 remarkable (value) \\
        11 catalytic performance (property) \\
        12 wide (value) \\
        13 pore size distribution (property) \\
        14 composite carbon supports (support) \\
        15 Pt-integrated catalyst (catalyst) \\
        16 outstanding (value) \\
        17 electrochemical performance (property) \\
        18 mass activity (property) \\
        19 8.6 times higher (value) \\
        20 commercial Pt/C catalyst (catalyst) \\
        21 hybrid carbon support (support) \\
        22 significantly enhance (value) \\
        23 durability (property) \\
        24 catalyst (material\_reference) \\
        25 high (value) \\
        26 power density (property) \\
        27 1.83 W cm$^{-2}$ (value) \\
        28 4 A cm$^{-2}$ (condition) \\
        \\
        \\
        \\
        \\
        \end{tabular}} 
        & 
        \resizebox{0.45\textwidth}{!}{
        \begin{tabular}[c]{@{}l@{}}
        1 (platinum-based electrocatalysts, \\proton exchange membrane, related\_to) \\
        2 (PtCo, simple synthesis strategy, \\related\_to) \\
        3 (PtCo, ultrafine, related\_to) \\
        4 (PtCo, alloy, related\_to) \\
        5 (PtCo, nanoparticles, related\_to) \\
        6 (PtCo, carbon support, related\_to) \\
        7 (PtCo, zeolitic imidazolate framework-67 \\(ZIF-67), related\_to) \\
        8 (PtCo, Ketjen Black (KB), related\_to) \\
        9 (PtCo, catalytic performance, related\_to) \\
        10 (catalytic performance, remarkable, \\related\_to) \\
        11 (pore size distribution, wide, related\_to) \\
        12 (composite carbon supports, \\pore size distribution, related\_to) \\
        13 (Pt-integrated catalyst, electrochemical \\performance, related\_to) \\
        14 (Pt-integrated catalyst, mass activity, \\related\_to) \\
        15 (electrochemical performance, \\outstanding, related\_to) \\
        16 (mass activity, 8.6 times higher, \\related\_to) \\
        17 (hybrid carbon support, durability, \\related\_to) \\
        18 (catalyst, power density, related\_to) \\
        19 (power density, high, related\_to) \\
        20 (power density, 1.83 W cm$^{-2}$, related\_to) \\
        21 (power density, 4 A cm$^{-2}$, related\_to) \\
        22 (1.83 W cm$^{-2}$, 4 A cm$^{-2}$, related\_to) \\
        \end{tabular}} \\ \hline
        \end{tabular}
        \end{table}

    \item \textbf{Brat Visualization} (Figure~\ref{fig:brat_output}):  Presents a structured annotation view, where extracted entities and relationships are visually highlighted. This format allows users to easily verify entity-relationship mappings
        \begin{figure}[!ht]
        \centering 
        \includegraphics[width=12cm, height=8cm]{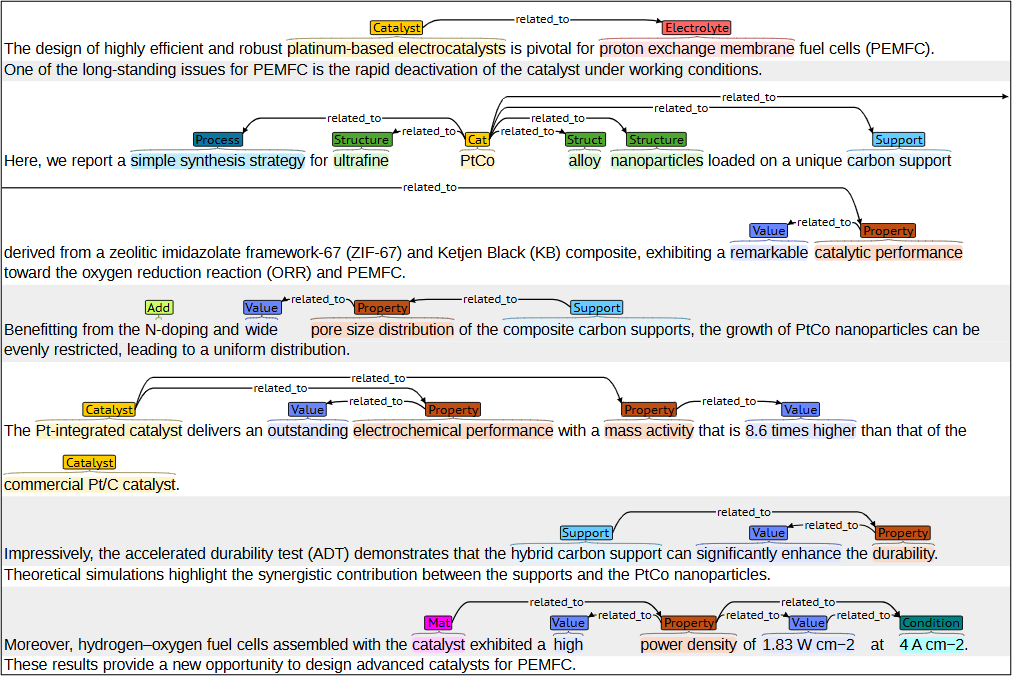}
        \caption{An example of the extracted data in Brat visualization for the text data taken from \cite{Zhang2024-1}.}
        \vspace{-6mm}
        \label{fig:brat_output}
        \end{figure}
    
    \item \textbf{Graph Visualization} (Figure~\ref{fig:graph_output}): Represents the extracted data as an interactive graph using PyVis, a library for constructing and visualizing network graphs. This dynamic visualization enhances interpretability by illustrating connections between entities in a structured manner
        \begin{figure}[!ht]
        \centering
        \includegraphics[width=7.3cm, height=6.5cm]{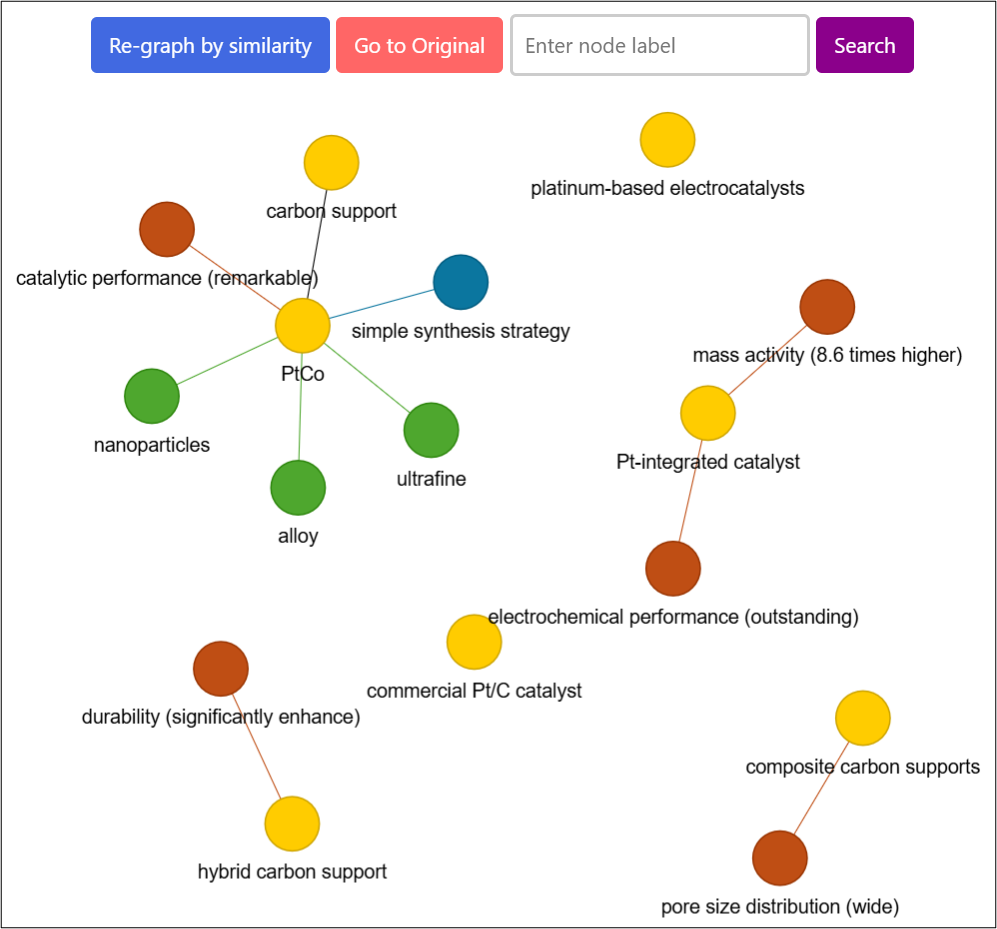}
        \caption{Extracted data in graph visualization.}
        \vspace{-6mm}
        \label{fig:graph_output}
        \end{figure}
\end{enumerate}

\section{Results and Discussion}
\label{result_and_discussion}
In this section, we present modeling processes, evaluating the performance of our trained models, and the outcomes of our data extraction on ORR catalyst-related scientific literature. We analyze the accuracy, precision, recall, and F1-score of different BERT-based models to determine their effectiveness in recognizing entities and relationships.

\subsection{Experiment Setup}
We annotated 76 articles, resulting in a total of 554 documents in the fuel cell dataset (Table~\ref{tab:statistics}) after converting them into a DyGIE++ compatible format. The dataset was split into training ($80$\%), validation ($10$\%), and test ($10$\%) sets. 

\begin{table}[h]
\centering 
\caption{Dataset statistics.}
\vspace{-3mm}
\label{tab:statistics}
\resizebox{0.50\textwidth}{!}{ 
\begin{tblr}{
  hline{1-2,8} = {-}{},
}
\textbf{Metric}                & \textbf{Value} \\
Number of articles             & 76             \\
Number of documents in DyGIE++ & 554            \\
Number of sentences            & 8,898          \\
Average sentences per article  & 117.07         \\
Number of annotated entities   & 16,301         \\
Number of annotated relations  & 13,899         
\end{tblr}
}
\end{table}

We fine-tuned \emph{seven} different pre-trained BERT-based models (1. SciBERT, 2. MatSciBERT-1, 3. MatSciBERT-2, 4. MatSciBERT-3, 5. PubMedBERT, 6. BlueBERT, and 7. BioBERT) on the fuel cell dataset, generating \emph{seven} specialized models. The three MatSciBERT variants differ in their pre-training or fine-tuning objectives: MatSciBERT-1 (MatSciBERT \cite{matscibert}) is the general-purpose model pre-trained on materials science papers from ScienceDirect; MatSciBERT-2 (MatSciBERT-CNER \cite{matscibert_cner}) is fine-tuned for Chemical Named Entity Recognition (CNER), enhancing its ability to identify chemical entities; and MatSciBERT-3 (MatScicBERT-Finetuned-SQuAD-PyTorch \cite{matscibert_squad}) is fine-tuned on the SQuAD dataset using PyTorch, improving its performance in span-based question answering tasks within scientific texts.

In addition to the standard \emph{Test} set, we included a \emph{Gold Standard} dataset, which consists of expert-verified annotations, as an additional evaluation set to gain deeper insights into the performance and characteristics of the fuel-cell trained models.

\subsection{NER and RE Models Performance}

We evaluated the performance of NER and RE models using DyGIE++, employing standard metrics such as Precision, Recall, and F1-score. The F1-scores were analyzed across four stages: \emph{Training}, \emph{Validation}, \emph{Testing}, and \emph{Gold Standard}, as shown in Table~\ref{tab:separate_models}.  

Overall, the NER F1-score ranged from 61.66\% to 82.19\%, while the RE F1-score ranged from 51.27\% to 66.10\%.

\begin{table}[h]
\centering
\caption{Performance of fuel-cell trained models by DyGIE++.}
\label{tab:separate_models}
\resizebox{1.10\textwidth}{!}{ 
\begin{tabular}{clcccccccc} 
\hline
\multirow{3}{*}{\begin{tabular}[c]{@{}c@{}}\textbf{Fuel-cell }\\\textbf{Model}\end{tabular}} & \multicolumn{1}{c}{\multirow{3}{*}{\begin{tabular}[c]{@{}c@{}}\textbf{Finetune }\\\textbf{Pre-trained}\\\textbf{ Model}\end{tabular}}} & \multicolumn{8}{c}{\textbf{F1-score}}                                                                                                                                                                                                                                                                                                                                                                            \\ 
\hhline{~~--------}
                                                                                             & \multicolumn{1}{c}{}                                                                                                                   & \multicolumn{2}{c}{{\cellcolor[rgb]{0.792,0.933,0.984}}\textbf{Training}}                          & \multicolumn{2}{c}{{\cellcolor[rgb]{0.851,0.949,0.816}}\textbf{Validation}}                        & \multicolumn{2}{c}{{\cellcolor[rgb]{0.984,0.89,0.839}}\textbf{Testing}}                          & \multicolumn{2}{c}{{\cellcolor[rgb]{0.796,0.808,0.984}}\textbf{Gold Standard}}                      \\ 
\hhline{~~--------}
                                                                                             & \multicolumn{1}{c}{}                                                                                                                   & {\cellcolor[rgb]{0.792,0.933,0.984}}\textbf{NER} & {\cellcolor[rgb]{0.792,0.933,0.984}}\textbf{RE} & {\cellcolor[rgb]{0.851,0.949,0.816}}\textbf{NER} & {\cellcolor[rgb]{0.851,0.949,0.816}}\textbf{RE} & {\cellcolor[rgb]{0.984,0.89,0.839}}\textbf{NER} & {\cellcolor[rgb]{0.984,0.89,0.839}}\textbf{RE} & {\cellcolor[rgb]{0.796,0.808,0.984}}\textbf{NER} & {\cellcolor[rgb]{0.796,0.808,0.984}}\textbf{RE}  \\ 
\hline
1                                                                                            & SciBERT                                                                                                                                & 99.76\%                                          & 98.40\%                                         & 64.31\%                                          & 53.11\%                                         & 61.36\%                                         & 48.93\%                                        & \textbf{82.19\%}                                 & 64.40\%                                          \\
2                                                                                            & MatSciBERT-1                                                                                                                           & 99.74\%                                          & 98.28\%                                         & 63.81\%                                          & 53.29\%                                         & \textbf{61.66\%}                                & \textbf{51.27\%}                               & 81.69\%                                          & \textbf{66.10\%}                                 \\
3                                                                                            & MatSciBERT-2                                                                                                                           & 99.67\%                                          & 98.34\%                                         & 65.84\%                                          & 54.74\%                                         & 59.01\%                                         & 45.86\%                                        & 79.41\%                                          & 61.94\%                                          \\
4                                                                                            & MatSciBERT-3                                                                                                                           & 99.73\%                                          & 98.40\%                                         & 65.63\%                                          & 54.18\%                                         & 61.02\%                                         & 50.36\%                                        & 80.57\%                                          & 66.08\%                                          \\
5                                                                                            & PubMedBERT                                                                                                                             & 99.71\%                                          & 98.28\%                                         & \textbf{67.30\%}                                 & \textbf{55.85\%}                                & 61.06\%                                         & 48.66\%                                        & \textbf{82.19\%}                                 & 64.95\%                                          \\
6                                                                                            & BlueBERT                                                                                                                               & 99.73\%                                          & 98.10\%                                         & 62.50\%                                          & 42.17\%                                         & 56.79\%                                         & 39.07\%                                        & 73.38\%                                          & 45.71\%                                          \\
7                                                                                            & BioBERT                                                                                                                                & 99.71\%                                          & 98.30\%                                         & 66.64\%                                          & 54.49\%                                         & 60.05\%                                         & 46.83\%                                        & 81.75\%                                          & 61.81\%                                          \\
\hline
\end{tabular}
}
\end{table}

\subsubsection{Performance of Models on Unseen Data}  
\begin{figure}[!ht]
\centering
\includegraphics[width=10.5cm, height=6.5cm]{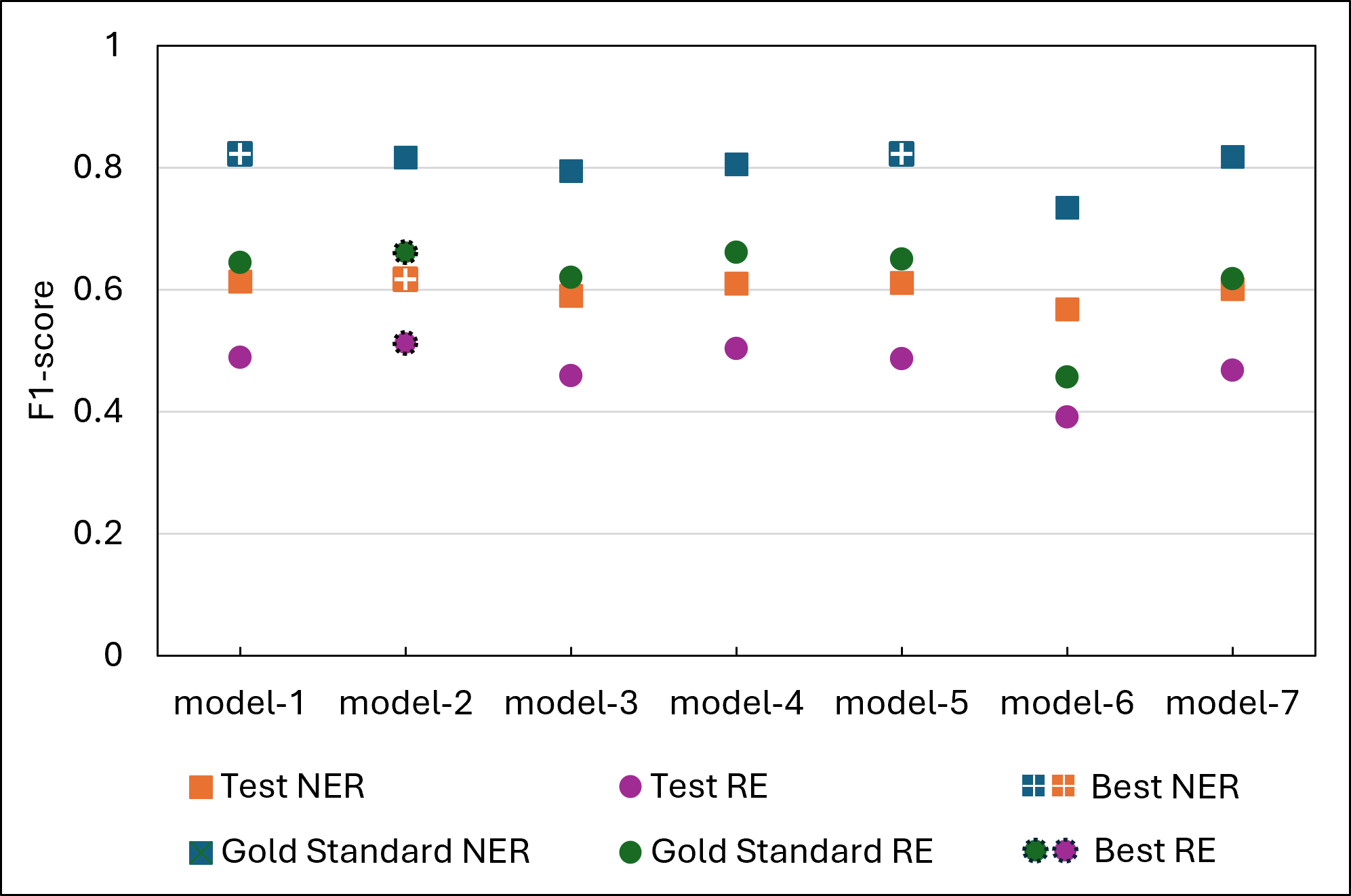}
\caption{Test and Gold Standard F1-scores.}
\label{fig:ner_re}
\end{figure}

To evaluate how well the fuel cell trained models generalize to unseen data, we focused on the F1-scores of the Test and Gold Standard sets, which reflect real-world performance. As shown in Figure~\ref{fig:ner_re}:  
\begin{itemize}
\setlength{\itemsep}{1pt}
\setlength{\parskip}{1pt}
    \item \textbf{On the Gold Standard test set}: Model-1 (SciBERT) and Model-5 (PubMedBERT) were the best-performing NER models, both achieving an F1-score of 82.19\%. Model-2 (MatSciBERT-1) achieved the highest RE F1-score of 66.10\%.
    \item \textbf{On the Test set}: Model-2 (MatSciBERT-1) outperformed all models in both NER and RE, achieving 61.66\% for NER and 51.27\% for RE.
\end{itemize}

These results indicate that PubMedBERT and SciBERT perform best in NER tasks on expert-verified data, while MatSciBERT-1 demonstrates robust performance across both NER and RE on real-world test data.

The models achieved higher NER and RE performance on the Gold Standard set. One possible reason is that the language and structure of the Gold Standard data, which was created by domain experts, offer higher consistency in sentence structure and clearer contextual cues. In contrast, the Test set comprises multiple texts and broader content diversity. Additionally, since the training data was annotated by non-experts who were trained using the Gold Standard annotations, the models may have indirectly learned annotation patterns aligned with the Gold Standard as a baseline, potentially giving them an advantage on this set.

\subsubsection{Overall Performance of Models}  
To establish a comprehensive ranking of the trained models, we considered the Validation F1-score alongside Test and Gold Standard scores. According to Table~\ref{tab:descend_order_models}, which presents the fuel cell trained models in descending order:  
\begin{itemize}
\setlength{\itemsep}{1pt}
\setlength{\parskip}{1pt}
    \item Model-5 (PubMedBERT) emerged as the top-performing model, followed by Model-2 (MatSciBERT-1), Model-4 (MatSciBERT-3), and Model-1 (SciBERT).  
    \item These models demonstrated the highest performance across all evaluation metrics.
    \item The remaining models showed relatively lower effectiveness in fuel cell related NLP tasks.
\end{itemize}

The ranking highlights that domain-specific BERT variants (e.g., PubMedBERT, MatSciBERT, and SciBERT) outperform general models like BlueBERT. This suggests that pre-trained models aligned with scientific and material science-related text are more suitable for fuel-cell NLP applications.

\begin{table}[h]
\centering
\caption{Fuel cell trained models in descending order.}
\label{tab:descend_order_models}
\resizebox{0.80\textwidth}{!}{ 
\begin{tabular}{ccc} 
\hline
\textbf{Models (↓ order)}       & \textbf{Fuel cell Model} & \textbf{Finetune Pre-trained Model}  \\ 
\hline
{\cellcolor[rgb]{0.314,0.867,0.867}}1 & model-5                  & PubMedBERT                           \\
{\cellcolor[rgb]{0.392,0.882,0.882}}2 & model-2                  & MatSciBERT-1                         \\
{\cellcolor[rgb]{0.471,0.898,0.894}}3 & model-4                  & MatSciBERT-3                         \\
{\cellcolor[rgb]{0.549,0.91,0.906}}4  & model-1                  & SciBERT                              \\
{\cellcolor[rgb]{0.631,0.925,0.918}}5 & model-7                  & BioBERT                              \\
{\cellcolor[rgb]{0.71,0.941,0.929}}6  & model-3                  & MatSciBERT-2                         \\
{\cellcolor[rgb]{0.788,0.953,0.945}}7 & model-6                  & BlueBERT                             \\
\hline
\end{tabular}
}
\end{table}

\subsection{Annotators vs. Model Performance}
We evaluated the performance of human annotators using standard metrics such as Precision, Recall, and F1-score, as described in Section~\ref{sec:modeling_process}. This evaluation was essential for establishing a benchmark against which model performance could be compared.

To ensure a fair comparison, we assessed both human annotators and model-generated extractions using the \emph{Gold Standard} dataset. This dataset consists of expert-verified annotations, making it the most reliable reference for evaluating extraction accuracy between annotators and models. 

For the \emph{annotator evaluation}, three annotators were provided with the same article used to create the Gold Standard dataset, and their performance was measured using the defined metrics. 

For the \emph{model evaluation}, we selected the top three trained models with the highest F1-scores and applied them to the same article. Their performance was evaluated based on the metrics used in DyGIE++.

\begin{figure}[!ht]
\centering
        \includegraphics[width=10.5cm, height=5.5cm]{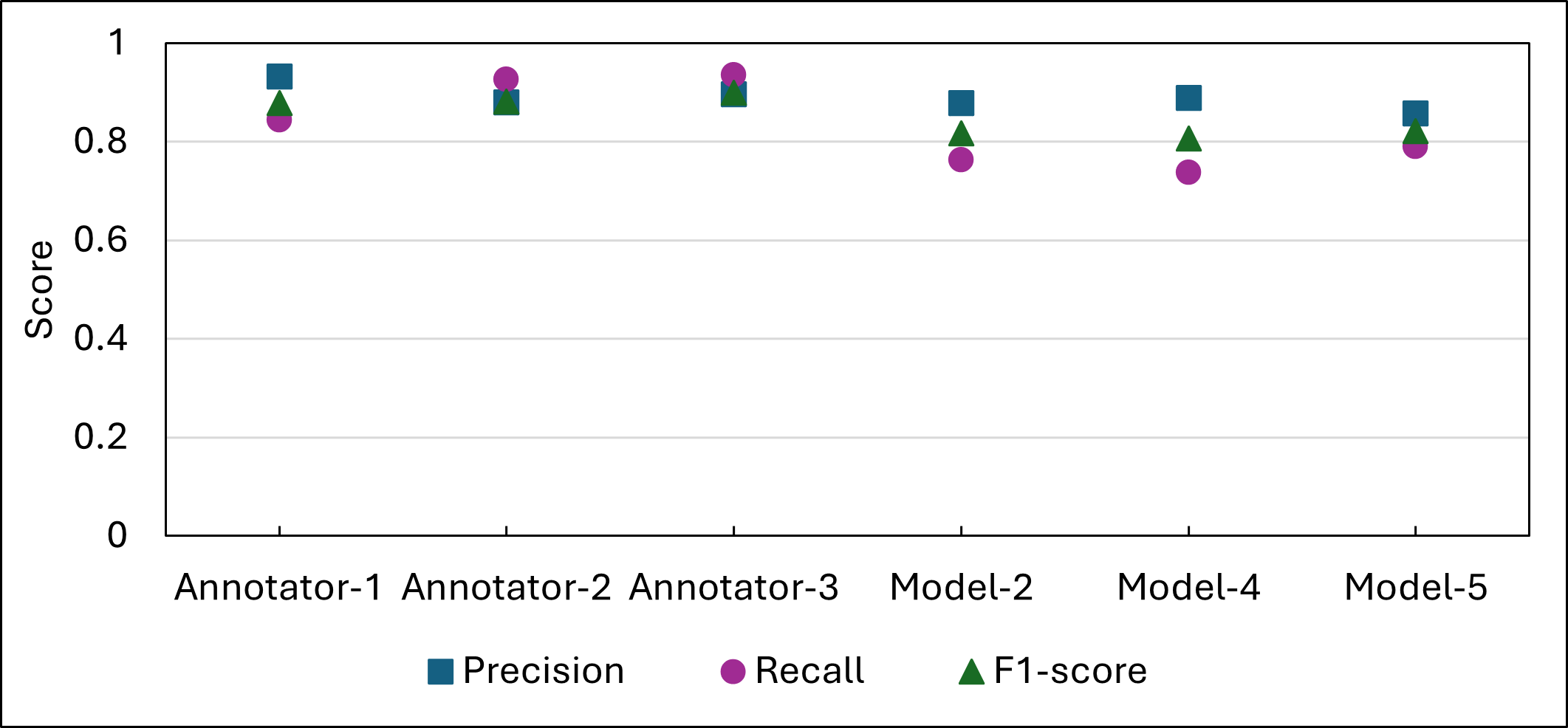}
        \caption{NER performance comparison}
        \label{fig:ner_performance}
\end{figure}

\begin{figure}[!ht]
\centering
        \includegraphics[width=10.5cm, height=5.5cm]{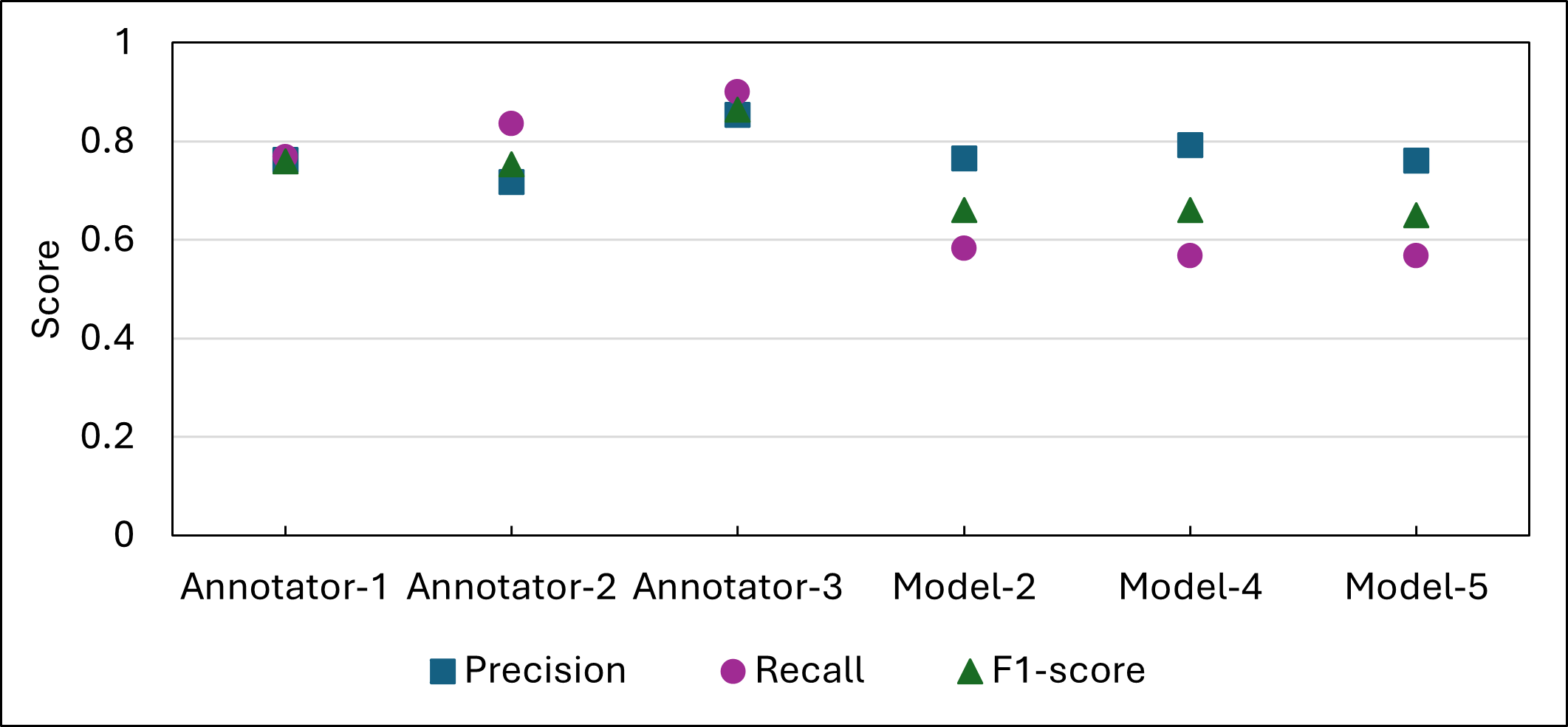}
        \caption{RE performance comparison}
        \label{fig:re_performance}
\end{figure}

This setup ensured that human and model performance were compared under identical conditions. Figure~\ref{fig:ner_performance} presents the NER performance comparison between annotators and models. Figure~\ref{fig:re_performance} illustrates the RE performance comparison. The performance gap in NER between annotators and models was not significant, indicating strong model reliability. However, for RE, models showed a noticeable drop in F1-score compared to annotators, suggesting that relation extraction remains more challenging for models.

The comparison results revealed the following trends:
\begin{itemize}
    \item \textbf{Annotators}: \textit{Precision} $<$ \textit{Recall}, indicating that human annotators tend to identify most relevant items but may also incorrectly label irrelevant entities and relationships.
    \item \textbf{Models}: \textit{Precision} $>$ \textit{Recall}, suggesting that models make more accurate predictions but may miss some relevant entities and relationships.
\end{itemize}
  
By evaluating both human annotators and models on the Gold Standard dataset, we ensured a consistent and objective performance assessment. The results suggest that models achieved comparable performance to human annotators for fuel cell literature extraction, particularly for NER. However, for RE, further improvements were necessary to enhance model accuracy.

\subsection{Limitations and Future Work}
While the current models show promising results, there are several limitations that need to be addressed. First, the dataset used in this study is small and might not cover the full diversity of language and terminology found in scientific literature, especially in the field of fuel cells. This limitation could affect the generalization of the models to new and unseen data. Furthermore, the model struggles with complex cross-sentence relations, which can impact the accuracy of relation extraction.

Future work will aim to expand the dataset by incorporating a broader range of articles from diverse sources, ensuring a more comprehensive representation of the field. Moreover, we will collect and incorporate table and graph data from scientific articles, which often contain critical information, into the training dataset. 

\section{Conclusion}
\label{conclusion}
This study presents a hybrid approach combining manual annotation with automated machine learning techniques to extract critical information on ORR catalysts from scientific literature. Our methodology demonstrated effective performance in extracting relevant entities and their relationships, providing valuable insights into the materials and conditions that influence catalyst performance in fuel cells. The findings suggest that NLP techniques, particularly those based on transformer models like BERT, can significantly accelerate the discovery and development of advanced catalysts for energy applications.

The insights derived from this study have significant implications for catalyst research. By automating the process of literature mining, researchers can quickly identify promising materials and synthesis techniques, reducing the time and effort required for manual analysis. This could lead to faster innovation and the development of more efficient and sustainable catalysts for ORR in fuel cells, contributing to advancements in clean energy technologies.

Future research could focus on extending the dataset to include a wider range of fuel cell technologies and catalyst types. Additionally, integrating multi-modal data sources, such as experimental data and computational models, could provide a more comprehensive understanding of catalyst behavior. Lastly, further improvements in relation extraction models could enable the prediction of new catalyst properties and performance based on extracted knowledge, ultimately advancing the field of energy materials science.

\section*{Appendix A: Supplementary Information}
\label{supplementary_data}
The extracted ORR catalyst data from 76 full-text articles are publicly available in CSV format. The dataset consists of 18 columns, including line\#ID, catalyst, support, additive, electrolyte, precursors, other\_material, material\_reference, property, property\_value, structure, structure\_value, process, process\_value, condition, condition\_value, and related\_material’s\_line\#ID. Each row represents a material along with its associated properties, while related materials are linked through the related\_material’s\_line\#ID.

The supplementary dataset can be accessed at: \href{https://thersacjp-my.sharepoint.com/:x:/g/personal/yp_19y_9613_f_thers_ac_jp/EdW66vtbfsFNmAChC1gD80ABhUb3KpH_pwKl-y5I6iT7XA?email=asahi.ryoji.d9%40f.mail.nagoya-u.ac.jp&e=MoWe7w}{Extracted ORR Catalyst Information (CSV)}.

\section*{CRediT authorship contribution statement}
\label{authorship_contribution}
\textbf{Hein Htet:} Conceptualization, Methodology, Software, Resources, Data Curation, Writing - Original Draft, Visualization.
\textbf{Amgad Ahmed Ali Ibrahim:} Conceptualization, Resources, Data Curation, Writing - Review \& Editing, Supervision.
\textbf{Yutaka Sasaki:} Conceptualization, Methodology, Resources, Writing - Review \& Editing, Supervision.
\textbf{Ryoji Asahi:} Conceptualization, Methodology, Resources, Data Curation, Writing - Review \& Editing, Supervision, Project administration.

\section*{Declaration of competing interest}
\label{conflict_interest}
The authors declare that they have no known competing financial interests or personal relationships that could have appeared to influence the work reported in this paper.

\section*{Acknowledgments}
\label{acknowledgement}
This paper is based on results obtained from a project, JPNP20003, commissioned by the New Energy and Industrial Technology Development Organization (NEDO).

\section*{Data availability}
\label{data_availability}
The data is publicly available on Mendeley Data as described in the article.


\end{document}